\documentclass{article} % For LaTeX2e
\usepackage{conference,times}

\usepackage{amsmath,amsfonts,bm}

\def\eqref#1{equation~\ref{#1}}
\def\1{\bm{1}}

\DeclareMathAlphabet{\mathsfit}{\encodingdefault}{\sfdefault}{m}{sl}
\SetMathAlphabet{\mathsfit}{bold}{\encodingdefault}{\sfdefault}{bx}{n}

\usepackage{hyperref}
\usepackage{url}
\usepackage{graphicx}
\usepackage{subcaption}
\usepackage{booktabs} 
\usepackage{caption}
\usepackage{geometry}
\usepackage{multirow}
\usepackage{cleveref}
\title{ABench-Physics: Benchmarking Physical Reasoning in LLMs via High-Difficulty and Dynamic Physics Problems}

\author{
\makebox[\textwidth][c]{%
  \parbox{0.95\textwidth}{\centering
Yiming Zhang\textsuperscript{1,2}, 
Yingfan Ma\textsuperscript{2}, 
Yanmei Gu\textsuperscript{2}, 
Zhengkai Yang\textsuperscript{2},\\
Yihong Zhuang\textsuperscript{2},
Feng Wang\textsuperscript{2}, 
Zenan Huang\textsuperscript{2}, 
Yuanyuan Wang\textsuperscript{2}, \\
Chao Huang\textsuperscript{2}, 
Bowen Song\textsuperscript{2}, 
Cheng Lin\textsuperscript{2}, 
Junbo Zhao\textsuperscript{1,2}
  }
} \\[0.5em]
\makebox[\textwidth][c]{%
  \parbox{0.8\textwidth}{\centering
    \textsuperscript{1} Zhejiang University\\
    \textsuperscript{2} Ant Group
  }
} \\[0.5em]
\makebox[\textwidth][c]{%
  \parbox{0.8\textwidth}{\centering
    \texttt{yimingz@zju.edu.cn}, 
    \texttt{j.zhao@zju.edu.cn}
  }
}
}

\newcommand{\phyA}{\textbf{\texttt{Phy\_A}} }
\newcommand{\phyB}{\textbf{\texttt{Phy\_B}} }

\iclrfinalcopy % Uncomment for camera-ready version, but NOT for submission.
\begin{document}

\maketitle

\begin{abstract}
Large Language Models (LLMs) have shown impressive performance in domains such as mathematics and programming, yet their capabilities in physics remain underexplored and poorly understood. Physics poses unique challenges that demand not only precise computation but also deep conceptual understanding and physical modeling skills. Existing benchmarks often fall short due to limited difficulty, multiple-choice formats, and static evaluation settings that fail to capture physical modeling ability. In this paper, we introduce ABench-Physics\footnote{\url{https://github.com/inclusionAI/ABench/tree/main/Physics}}, a novel benchmark designed to rigorously evaluate LLMs’ physical reasoning and generalization capabilities. ABench-Physics consists of two components: \phyA, a static set of 400 graduate- or Olympiad-level problems; and \phyB, a dynamic subset of 100 problems equipped with an automatic variation engine to test model robustness across changing conditions. All questions require precise numerical answers, with strict formatting and tolerance constraints. Our evaluation of several state-of-the-art LLMs reveals substantial performance gaps, highlighting persistent limitations in physical reasoning, especially in generalization to dynamic variants. ABench-Physics provides a challenging and diagnostic framework for advancing scientific reasoning in LLMs.
\end{abstract}

\section{Introduction}

In recent years, Large Language Models (LLMs) have demonstrated remarkable progress across various domains, including mathematics \citep{gao2024omnimath,xu2025ugmathbench,kydlicek2021mathverify,liu2024mathbench}, programming \citep{jain2024livecodebench,ahmad2025opencodereasoning}, common-sense reasoning \citep{liu2020logiqachallengedatasetmachine,han-etal-2024-folio} and others \citep{huang2024c}. However, physics \citep{planck1949scientific, hawking2009brief}, the bedrock of all natural sciences, presents a unique set of challenges that extend beyond mere computational prowess. Solving physics problems requires not only precise calculation but also a deep understanding of abstract concepts, the ability to construct physical models \citep{chung2025theoreticalphysicsbenchmarktpbench, zheng2025scalingphysicalreasoningphysics, wang2024scibench}, and the skill to apply fundamental principles in complex scenarios.

Although the research community has released several benchmarks to evaluate the physics capabilities of LLMs, they generally suffer from several core limitations \citep{qiu2025phybench,xu2025ugphysics}. First, many existing benchmarks (e.g., MMLU-Physics \citep{li2024cmmlu}) focus on high school or introductory university-level problems and are often presented in a multiple-choice format, which fails to comprehensively assess a model's ability to generate complex solution paths and precise numerical answers. Second, while introduce higher-difficulty problems, they often face the risk of data contamination\citep{deng2024unveilingspectrumdatacontamination,chen2025recentadvanceslargelangauge}. Models may have seen similar problems in their vast training data, leading to inflated performance metrics that do not reflect physical modeling abilities. Most importantly, existing benchmarks are almost entirely static. This means a model can achieve a high score by memorizing solution patterns for specific problems rather than by genuinely mastering versatile physical reasoning skills \citep{xie2025memorizationlargelanguagemodels}. Such an evaluation paradigm cannot effectively verify if a model can generalize its knowledge to solve novel problems with slight variations in conditions.

\begin{figure}[ht]
  \centering
  \begin{subfigure}[t]{0.48\linewidth}
    \centering
    \includegraphics[width=\linewidth]{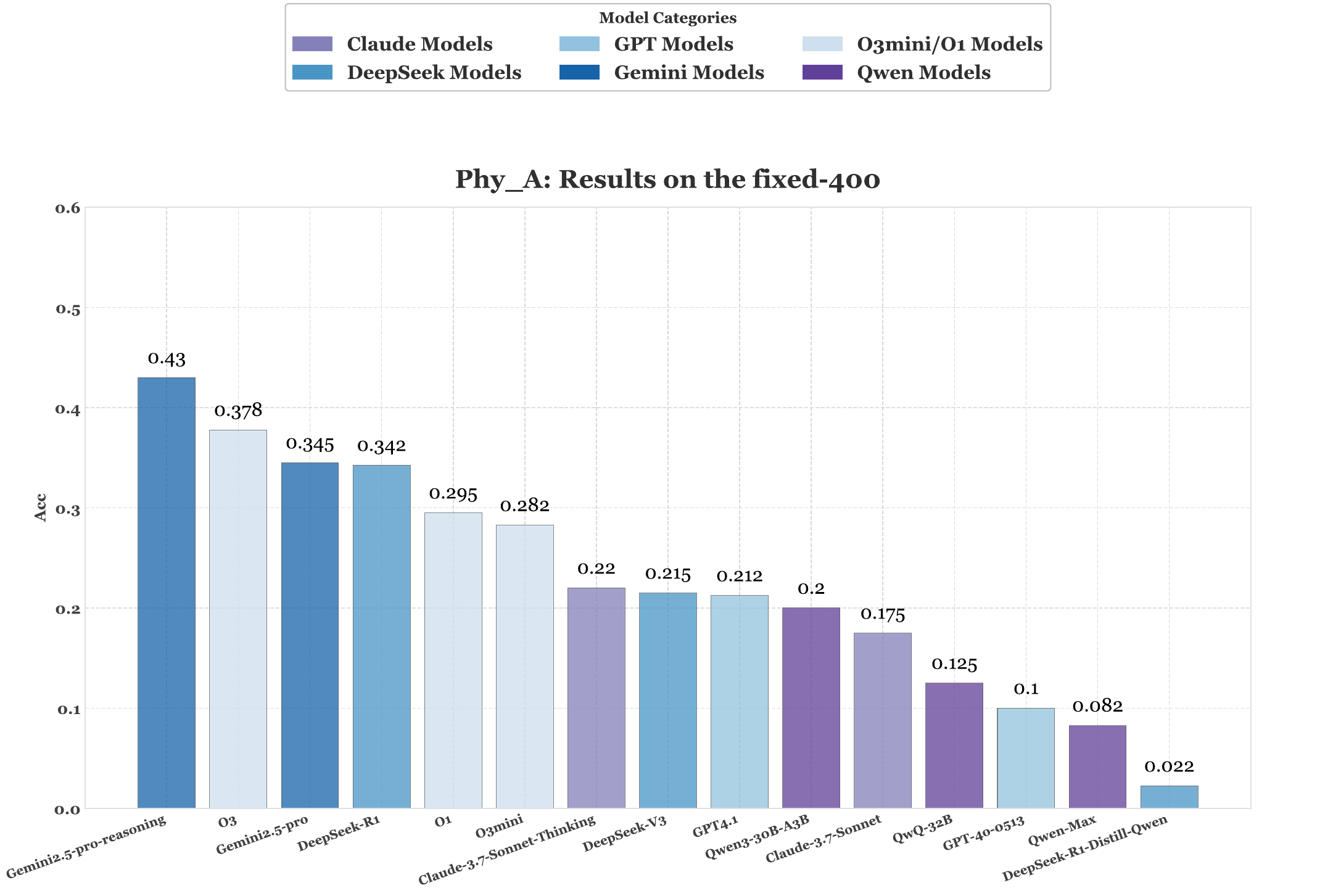}
    \caption{Accuracy on the static \texttt{Phy\_A\_fixed\_400} subset.}
    \label{fig:phy_a}
  \end{subfigure}
  \hfill
  \begin{subfigure}[t]{0.48\linewidth}
    \centering
    \includegraphics[width=\linewidth]{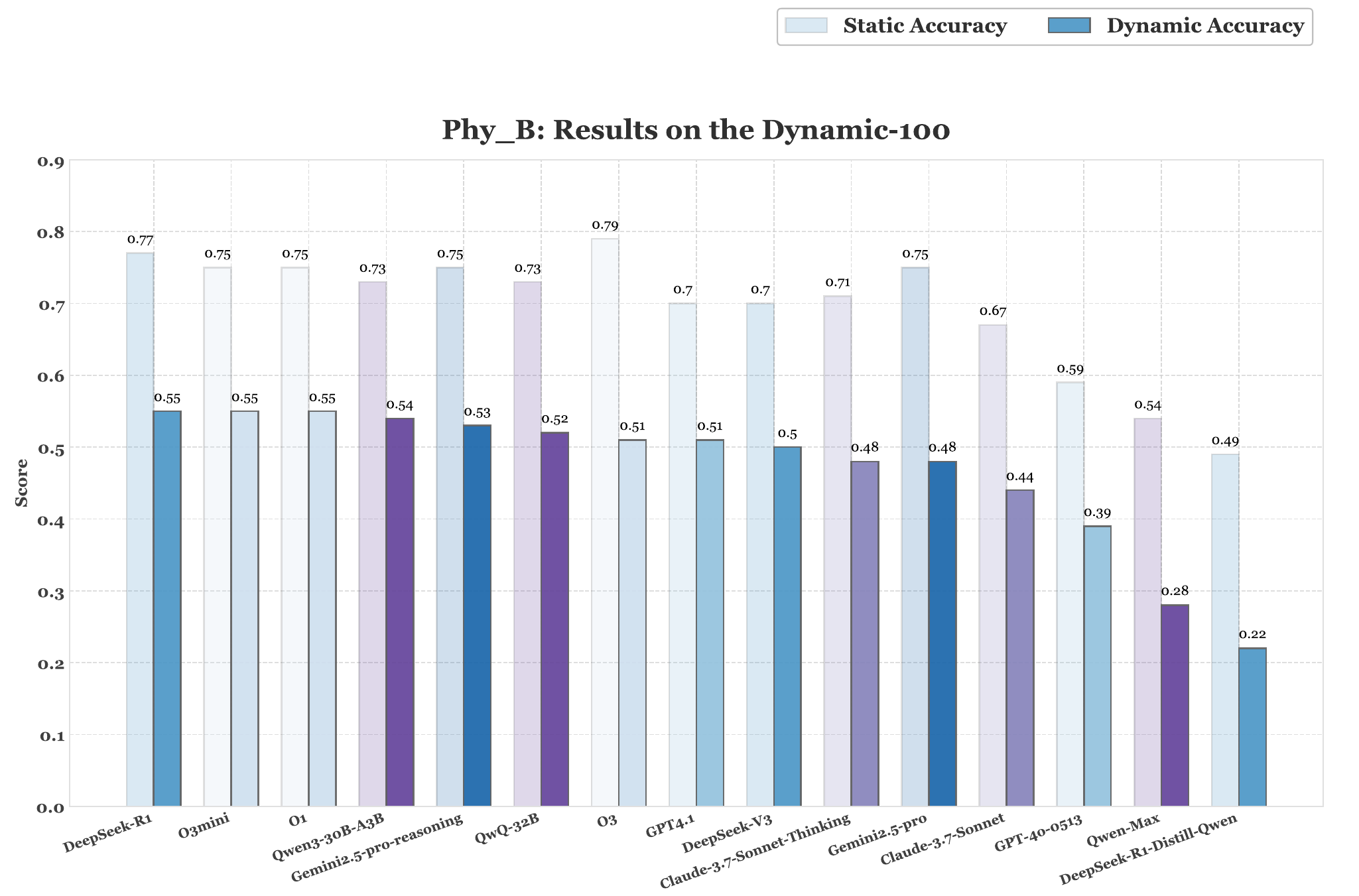}
    \caption{Accuracy on the dynamic \texttt{Phy\_B\_dynamic\_100} subset.}
    \label{fig:phy_b}
  \end{subfigure}
  \caption{Model accuracy on the two benchmark subsets.}
  \label{fig:benchmark_accuracy}
\end{figure}

Therefore, we introduce \textbf{ABench-Physics}, a novel benchmark designed to establish a rigorous evaluation framework that specifically measures LLMs' genuine reasoning capabilities in physics. 
Our evaluation focuses on both the model’s ability to reason through complex and difficult physics problems, and its physical modeling capacity to generalize across physical formulations.
Specifically, \textbf{ABench-Physics} consists of two unique parts. 
\phyA (\texttt{Phy\_A\_fixed\_400}) is a fixed set of 400 high-difficulty, static graduate-level or Olympiad-level physics problems, providing a stable and reproducible baseline for model performance. 
\phyB (\texttt{Phy\_B\_dynamic\_100}) is a collection of 100 dynamic physics problems, whose core innovation is an automatic generation engine that alters numerical values within the core LaTeX equations to produce a vast number of variants with identical physical models but different conditions. This design compels the model to demonstrate genuine physical modeling and computational ability, rather than mere rote memorization.

Unlike previous benchmarks that adopted multiple-choice formats~\citep{li2024cmmlu}, which reduce question difficulty, or required answers in the form of expressions~\citep{qiu2025phybench}, which are difficult to evaluate precisely, our benchmark focuses exclusively on numerical calculation problems.
We evaluate numerical answers with a strict 1\% tolerance for error. For the dynamic \phyB set, we employ an even stricter metric: a model is only credited with a correct answer if it successfully solves all variations of a given problem, which stringently tests the model's adaptability and generalization capabilities.
To eliminate discrepancies caused by answer formatting, each question explicitly specifies the required unit and the number of significant figures. Additionally, we provide a lightweight LaTeX-based evaluation function that can handle various numerical formats of the predicted answer.
 
As shown in \Cref{fig:benchmark_accuracy}, we conducted a comprehensive evaluation of a series of state-of-the-art LLMs on \textbf{ABench-Physics}. Our results clearly indicate that even the most advanced models, such as Gemini 2.5 Pro, perform poorly on our static, high-difficulty problems (\phyA), achieving a top accuracy of only 43.0\%. This confirms that a significant gap remains in advanced physics reasoning for current LLMs. More critically, when transitioning from static to dynamic problems (\phyB), all models exhibited a substantial average performance drop of 22.5\%. This finding powerfully demonstrates the effectiveness of \textbf{ABench-Physics} in revealing models' reliance on memorization. We argue that \textbf{ABench-Physics} provides the community with a novel and rigorous tool to drive the development of LLMs from simple pattern matching towards deeper and more robust scientific reasoning. 

\section{Related Work}
\textbf{Datasets for Physical Reasoning}
The application of large language models (LLMs) in scientific domains has led to the development of a variety of benchmarks to assess their physical reasoning capabilities \citep{jaiswal2024improving, arora2023have}, moving from foundational multiple-choice evaluations to more complex and demanding frameworks \citep{qiu2025phybenchholisticevaluationphysical,bakhtin2019phyrenewbenchmarkphysical, chung2025theoreticalphysicsbenchmarktpbench, ding2023using}. Recent efforts have significantly expanded the scope of text-based physical reasoning evaluation in both depth and scale. Benchmarks like UGPhysics \citep{xu2025ugphysicscomprehensivebenchmarkundergraduate} have significantly advanced the evaluation depth by introducing challenging, university-level physics problems and developing sophisticated evaluation protocols to handle diverse symbolic and numerical answer formats. Concurrently, initiatives such as the PHYSICS dataset \citep{zheng2025scalingphysicalreasoningphysics} have focused on achieving unprecedented scale and trainability, providing the largest available corpus of physics problems explicitly split into training and test sets to facilitate model improvement through fine-tuning. A common finding from these works is that even state-of-the-art models exhibit significant limitations in complex physics reasoning, often pointing to deficits in foundational knowledge. Parallel to these text-based advancements \citep{hendrycks2020measuring}, the scope of evaluation has expanded into a new dimension with the introduction of benchmarks like PhysBench \citep{chow2025physbench}. This work pioneers the assessment of visual physical reasoning in Vision-Language Models (VLMs) by using entries of interleaved video, images, and text, evaluating a model's intuitive grasp of physical phenomena, such as object dynamics and properties. 

\textbf{Limitations of Static Evaluation Paradigms}
Despite the significant advances across different axes, depth, scale, and modality, a fundamental limitation persists across these evaluation paradigms: their reliance on static test sets. This characteristic makes them inherently vulnerable to data contamination, where models can achieve high scores by retrieving memorized answers rather than engaging in genuine reasoning\citep{deng2024unveilingspectrumdatacontamination,GAO2025130135, zhang2025puzzlebenchfullydynamicevaluation,li2024dynaeval}. The static nature of existing benchmarks makes it difficult to decouple memorization from the ability to generalize knowledge to novel scenarios \citep{xie2025memorizationlargelanguagemodels}. This highlights a critical gap in the field: the need for an evaluation framework that can dynamically assess a model's reasoning robustness and generalization \citep{kiyomaru-etal-2024-comprehensive-analysis, lu2024scalinglawsfactmemorization}. Such a framework must be able to verify whether a model has truly mastered physical principles or has simply overfitted to the specific problem distributions seen during training.

% \section{ABench-Physics Dataset}
\section{ABench-Physics Overview}

\subsection{Dataset Composition and Key Properties}
\label{subsec:dataset}
The released corpus—hereafter \textbf{Physics Benchmark}—contains \textbf{500 carefully verified problems} that span the breadth of an advanced undergraduate curriculum and popular competition syllabi, namely classical mechanics, electromagnetism, optics, quantum physics, electrodynamics, semiconductor and fluid mechanics, atomic physics, and assorted modern topics.  
To support both reasoning ability and physical modeling capacity evaluation, the material is organized into two complementary parts.  
\phyA contributes \textbf{400 high-difficulty, static questions}. This subset of questions is primarily selected from graduate-level or Olympiad-level physics problems, based on experimental screening that identified cases where models tend to produce long responses with low accuracy.
\phyB offers \textbf{100 parameterised questions} with an accompanying parameter variation engine that regenerates fresh variants by perturbing the numerical constants embedded in the ground-truth \LaTeX{} solutions; this dynamic design thwarts memorisation and measures a model’s physical modeling adaptability. \footnote{The development of the engine is still ongoing. To facilitate early collaboration and feedback from the research community, we release a preprint version of ABench-Physics with three variations for each question in \phyB. Future updates will expand the benchmark with additional content and improvements.}

\begin{table}[ht]
\centering
\caption{Composition of the Physics Benchmark.}
\label{tab:composition}
\begin{tabular}{lccc}
\toprule
Subset & \# Problems & Difficulty & Variability \\
\midrule
\phyA (static)   & 400 & High      & Fixed text \\
\phyB (dynamic)  & 100 & Moderate  & Parameterised via numeric perturbation \\
\bottomrule
\end{tabular}
\end{table}

\subsection{Construction Pipeline}
\label{subsec:construction}
To construct our benchmark dataset, we began by collecting physics problems from a wide range of sources, including textbooks and examination papers. The selected questions were digitized using optical character recognition (OCR) tools and then manually extracted, followed by an automated quality check. Subsequently, each problem underwent a round of manual annotation to ensure structural and semantic accuracy.

Based on the annotated questions, we employed Deepseek-R1~\citep{deepseek2025deepseekr1} to rewrite the problem scenarios, introducing controlled variations in context. These rewrites were then manually reviewed and refined to guarantee correctness and consistency. To further ensure the quality of the dataset, every question—along with its corresponding solution—was independently verified by one to two human annotators, who attempted to solve the problem step by step based on the provided formulation.

Each question was paraphrased by an independent reviewer to minimize overlap with publicly available pre-training corpora. In addition, all symbolic derivations and numeric endpoints were independently produced and cross-verified. Each problem admits a unique and unambiguous answer—ensuring high-quality, reproducible evaluation.
We categorized the resulting questions into two subsets. More difficult problems that challenge model reasoning were selected as candidates for \phyA, while those with clearer computational procedures and well-defined intermediate steps were designated for \phyB.

\subsection{Evaluation Protocol}
\label{subsec:evaluation}
Since all questions in our benchmark are numerical in nature with a unique correct value, we adopt a straightforward evaluation protocol by directly comparing the model’s predicted answer with the ground truth, ensuring precise and objective assessment.
Given a model output, we first utilize our lightweight LaTeX-based evaluation function to translate the output into a numerical answer \(\hat{y}\). With \(\hat{y}\) and ground truth \(y^\star\), the answer is accepted when
\[
\left| \hat{y} - y^\star \right| \le 0.01\,\lvert y^\star\rvert,
\]
i.e.\ the relative error does not exceed \(1\%\).  There are still certain differences in the evaluation protocols between two subsets.
\emph{Fixed 400-Problem Accuracy} measures performance on the immutable \phyA subset, providing a stable yardstick across model iterations.  
\emph{Dynamic 100-Problem Accuracy} evaluates \phyB. The model must answer \emph{all} regenerated variants of a template correctly to receive credit, thereby probing robustness to numerical perturbations. 
Together, these scores capture both raw problem-solving competence and resilience to distributional shift.

\section{Experiments}
\label{sec:expconcl}

\begin{table}[htbp]
  \centering
  \caption{Comprehensive Evaluation of Models}
  \renewcommand{\arraystretch}{1.2}
  \begin{tabular}{l|cc|ccc}
    \toprule
    \multirow{2}{*}{\textbf{Models}} & \textbf{\phyA} & & \multicolumn{3}{c}{\textbf{\phyB}} \\
    \cmidrule(lr){2-3} \cmidrule(lr){4-6}
    & \textbf{Acc} & & \textbf{Stactic Acc} & \textbf{Dynamic Acc} & \textbf{$\Delta$} \\
    \midrule
    Gemini2.5-pro-reasoning\citep{google2025gemini}      & \textbf{0.430} & & 0.75 & 0.53 & 0.22 \\
    o3\citep{openai2025o3}                           & 0.378 & & 0.79 & 0.51 & \textbf{0.28} \\
    Gemini2.5-pro \citep{google2025gemini}               & 0.345 & & 0.75 & 0.48 & 0.27 \\
    DeepSeek-R1-0528\citep{deepseek2025deepseekr1}             & 0.343 & & 0.77 & \textbf{0.55} & 0.22 \\
    o1   \citep{openai2024o1}                     & 0.295 & & 0.75 & \textbf{0.55} & 0.20 \\
    o3mini\citep{openai2025o3}                         & 0.283 & & 0.75 & \textbf{0.55} & 0.20 \\
    Claude-3.7-Sonnet-Thinking   & 0.220 & & 0.71 & 0.48 & 0.23 \\
    DeepSeek-V3 \citep{deepseek2024deepseekv3}                 & 0.215 & & 0.70 & 0.50 & 0.20 \\
    GPT4.1 \citep{openai2025gpt41}                      & 0.213 & & 0.70 & 0.51 & 0.19 \\
    Qwen3-30B-A3B \citep{qwenlm2025qwen3}               & 0.200 & & 0.73 & 0.54 & 0.19 \\
    Claude-3.7-Sonnet \citep{anthropic2025claude}           & 0.175 & & 0.67 & 0.44 & 0.23 \\
    QwQ-32B \citep{qwen2025qwq32b}                     & 0.125 & & 0.73 & 0.52 & 0.21 \\
    GPT-4o-0513  \citep{openai2024gpt4o}                & 0.100 & & 0.59 & 0.39 & 0.20 \\
    Qwen-Max \citep{qwen2025qwen2.5}                    & 0.083 & & 0.54 & 0.28 & 0.26 \\
    DeepSeek-R1-Distill-Qwen-32B \citep{deepseek2025deepseekr1} & 0.023 & & 0.49 & 0.22 & 0.27 \\
    \bottomrule
  \end{tabular}
  \label{tab:Main-result}
\end{table}

\subsection{Main Result}
\Cref{tab:Main-result} presents the accuracy of four widely used large language models on the static \texttt{Phy\_A\_fixed\_400} set (\phyA) and the dynamic \texttt{Phy\_B\_dynamic\_100} set (\phyB). 

\textbf{Persistent Difficulty for state-of-the-art Models.}
Even the strongest baseline, Gemini-2.5 Pro \citep{google2025gemini}, solves only about 43\% of the static questions, and the remaining models cluster well below that mark.  The aggregate picture is clear: current SOTA LLMs still fail to reliably handle our physics challenge benchmark, underscoring a substantial gap between general language proficiency and rigorous quantitative reasoning.

\textbf{Sensitivity to Numerical Perturbation.}
Altering the constants embedded in each prompt produces a pronounced accuracy drop across the board.  Averaged over all systems, the decline is \textbf{22.5\%}. The most fragile baseline loses nearly \textbf{28\%}.  Because the semantic structure of every problem remains unchanged, these results indicate that the models often rely on superficial pattern matching rather than stable algebraic manipulation.

\textbf{RL Outperforms SFT on Dynamic Questions.}
A consistent ranking reversal occurs when moving from the static to the dynamic set: models fine-tuned with reinforcement learning from human or programmatic feedback register smaller relative losses and, in two cases, overtake instruction-tuned (SFT) counterparts that had previously outperformed others on \phyA.  This trend suggests that RL training better enables an LLM to handle out-of-distribution numerical variations, whereas pure supervised fine-tuning leaves the system brittle.

\section{Conclusion}

We present ABench-Physics, a new benchmark aimed at rigorously evaluating the genuine physics reasoning capabilities of LLMs. Unlike previous benchmarks, ABench-Physics focuses exclusively on numerical calculation problems, removes ambiguity in answer formats, and introduces dynamic variants to probe model generalization. Our two-part design, including static high-difficulty questions in \phyA and dynamically varying problems in \phyB, enables a fine-grained assessment of both reasoning depth and robustness to change. Empirical results across several leading LLMs reveal that even top-performing models struggle significantly with both static and dynamic physics tasks, exposing their limited ability to model and generalize physical concepts. We argue that ABench-Physics will serve as a valuable diagnostic tool for the community, encouraging the development of LLMs with stronger scientific understanding and reasoning skills.

\bibliography{conference}
\bibliographystyle{conference}

\end{document}